\documentclass{article}

\usepackage[numbers]{natbib}

    \usepackage[preprint]{neurips_2023}

\usepackage[utf8]{inputenc} 
\usepackage[T1]{fontenc}    
\usepackage{hyperref}       
\usepackage{url}            
\usepackage{booktabs}       
\usepackage{amsfonts}       
\usepackage{nicefrac}       
\usepackage{microtype}      
\usepackage{xcolor}         

\definecolor{darkgreen}{rgb}{0,0.5,0}
\newcommand{\cmark}{\textcolor{darkgreen}{\ding{51}}}
\newcommand{\xmark}{\textcolor{red}{\ding{55}}}
\usepackage{ulem}
\usepackage{multirow}
\usepackage{diagbox}
\usepackage{booktabs}
\usepackage{xcolor}
\definecolor{mygreen}{HTML}{2CB600}
\usepackage{subfig}
\usepackage{xspace}
\usepackage{appendix}
\usepackage{graphicx}
\usepackage{epsfig}
\usepackage{amsmath}
\usepackage{pifont}
\usepackage{tabularx}
\usepackage{seqsplit}
\usepackage{nameref}
\usepackage{varioref}
\usepackage{hyperref}
\usepackage{cleveref}
\usepackage{todonotes}
\usepackage{comment}

\usepackage{amssymb}
\usepackage{pifont}

\usepackage[section]{placeins}
\usepackage{wrapfig}

\newcommand{\resolved}[1]{}

\newcommand{\com}[1]{}

\newcommand{\draftcomment}[3]{{\textcolor{#3}{[#1]#2}}}
\renewcommand{\draftcomment}[3]{}  

\newcommand{\yonatan}[1]{\draftcomment{#1}{\textsc{yonatan}}{blue}}

\newcommand{\jack}[1]{\draftcomment{#1}{\textsc{jack}}{purple}}

\newcommand{\pw}[1]{\draftcomment{#1}{\textsc{pw}}{violet}}

\hypersetup{
  colorlinks   = true, 
  urlcolor     = blue, 
  linkcolor    = blue, 
  citecolor   = blue 
}

\newcommand{\benchmarkname}[0]{VisIT-Bench\xspace}
\newcommand{\sheetlink}{\href{https://docs.google.com/spreadsheets/d/1hi8rGXf2WYufkFvGJ2MZ92JNChliM1QEJwZxNboUFlE/edit?usp=sharing}{VisIT-Bench Sheet}}
\newcommand{\sheetlinkmulti}{\href{https://docs.google.com/spreadsheets/d/1IgCjJEd_obCawo1rWYfRZ_J7eiHP_68db5_OaNchKL0/edit?usp=sharing}{VisIT-Bench Sheet Multi-Images}}

\newcommand{\mailto}[1]{\href{mailto:#1}{\texttt{#1}}}

\newcommand{\elo}[0]{Elo\xspace}

\newcommand{\densecaption}[0]{instruction-conditioned caption\xspace}
\newcommand{\densecaptions}[0]{instruction-conditioned captions\xspace}

\newcommand{\customfootnotetext}[2]{{%
  \renewcommand{\thefootnote}{#1}%
  \footnotetext[0]{#2}}}%

\title{VisIT-Bench: A Benchmark for Vision-Language Instruction Following Inspired by Real-World Use}

\author{
\normalsize Yonatan Bitton*$^{1,2}$ 
\hspace{0.2em}
\normalsize Hritik Bansal*$^{3}$
\hspace{0.2em}
\normalsize Jack Hessel*$^{4}$
\hspace{0.2em}
\normalsize Rulin Shao$^{5}$
\hspace{0.2em}
\normalsize Wanrong Zhu$^{6}$ \\ 
\hspace{0.2em}
\normalsize \textbf{Anas Awadalla}$^{5}$
\hspace{0.2em}
\normalsize \textbf{Josh Gardner}$^{5}$
\hspace{0.2em}
\normalsize \textbf{Rohan Taori}$^{7}$
\hspace{0.2em}
\normalsize \textbf{Ludwig Schmidt}$^{4,5,8}$ \\ 
}

\begin{document}

\customfootnotetext{${^*}$}{Equal contribution. Contact {\mailto{yonatanbitton1@gmail.com},\mailto{hbansal@ucla.edu},\mailto{jackh@allenai.org},\mailto{ schmidt@cs.washington.edu}}. 
  $^{1}$Hebrew University \,
  $^{2}$Google Research \,
  $^{3}$UCLA \,
  $^{4}$Allen Institute for AI \,
  $^{5}$University of Washington \,
  $^{6}$UCSB \,
  $^{7}$Stanford \
  $^{8}$LAION \
  ,}

\maketitle

\begin{abstract}
We introduce \benchmarkname (\textbf{Vis}ual \textbf{I}ns\textbf{T}ruction \textbf{Bench}mark), a benchmark for evaluation of instruction-following vision-language models for real-world use. Our starting point is curating 70 ``instruction families'' that we envision instruction tuned vision-language models \textit{should} be able to address. Extending beyond evaluations like VQAv2 and COCO, tasks range from basic recognition to game playing and creative generation. Following curation, our dataset comprises 592 test queries, each with a human-authored \densecaption{}. These descriptions surface instruction-specific factors, e.g., for an instruction asking about the accessibility of a storefront for wheelchair users, the \densecaption describes ramps/potential obstacles. These descriptions enable 1) collecting human-verified reference outputs for each instance; and 2) automatic evaluation of candidate multimodal generations using a text-only LLM, aligning with human judgment. We quantify quality gaps between models and references using both human and automatic evaluations; e.g., the top-performing instruction-following model wins against the GPT-4 reference in just 27\% of the comparison.
\benchmarkname is dynamic to participate, practitioners simply submit their model's response on the project website; Data, code and leaderboard is available at \url{https://visit-bench.github.io/}.
\end{abstract}

\begin{figure}[h]
    \centering
    \includegraphics[width=\textwidth]{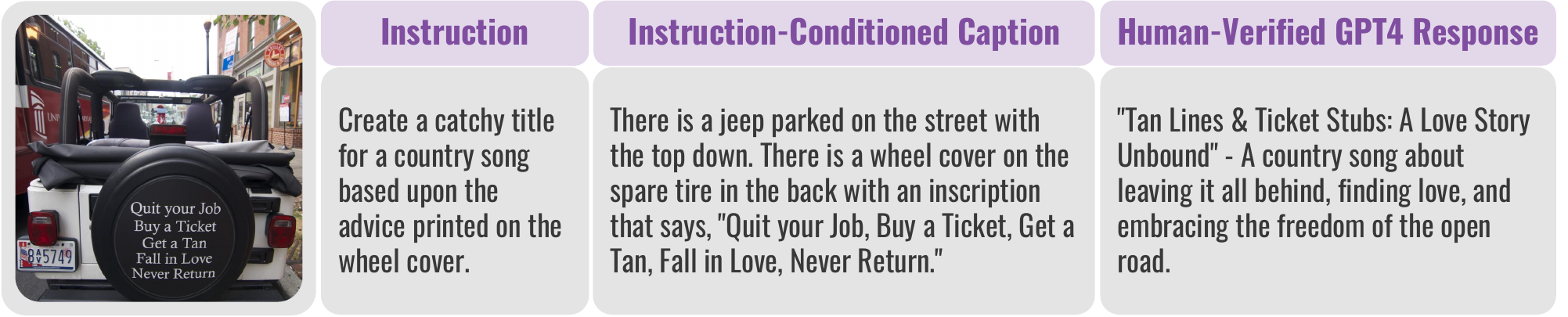}
    \caption{An example from \benchmarkname{}, featuring an image, a challenging instruction, an \densecaption{}, and a human-verified GPT4 response. These elements are used for evaluating multimodal chatbots and updating a dynamic leaderboard.}
    \label{fig:single_example}
\end{figure}

\section{Introduction}

A long-standing challenge for artificial intelligence is to build general-purpose assistants that can, in collaboration with humans, solve diverse and never-before-seen tasks \cite{askell2021general}.
For textual tasks, several recent works \cite{ouyang2022training,wei2021finetuned,wang2022super,alpaca,vicuna2023,gpt4} have shown that fine-tuning language models such as GPT-3 and LLaMA with supervised instruction+response examples  \cite{brown2020language,chowdhery2022palm,touvron2023llama} enables them to respond to imperative requests and questions without task-specific training. Zero-shot generalization is promising not only for standard academic benchmarks, but -- perhaps more-so -- for creative, useful, and real-world queries that downstream users of language technologies are likely to make.

On the multimodal side, recent instruction-following vision-language models also provide a zero-shot interface. Given an image (or multiple images) and a query (e.g., ``how many apples are in this image?'' or ``What is this?'' or ``Write a poem in the style of Robert Frost about this scene.'') a textual response is provided.
Recent works like OpenFlamingo \cite{anas_awadalla_2023_7733589,Alayrac2022FlamingoAV}, LLaVA \cite{liu2023llava} and others \cite{dai2023instructblip,li2023blip,gao2023llama, ye2023mplug, su2023pandagpt}, have implemented this interface with promising initial results. Although standard benchmarks like VQAv2 \cite{goyal2017making} and COCO captioning \cite{lin2014microsoft} are commonly used to assess performance, less is know about how models perform on broader, open-ended queries that resemble real-world user behavior. Evaluations of such queries typically rely on informal and qualitative approaches.

To support quantitative evaluation for this setting, we present \benchmarkname (\textbf{Vis}ual \textbf{I}ns\textbf{T}ruction \textbf{Bench}mark), a dynamic benchmark consisting of 592 challenging vision-language instructions. Each instance contains an instruction, input image(s), a \densecaption{}  (a human-crafted caption for the image(s)/instruction), and a human verified reference (\Cref{fig:single_example}).
Instructions are image-contextual imperative requests or questions, e.g., for an image of pancakes, a user asks \textit{``how can I cook this in a healthy way?''}.
Different from existing zero-shot evaluations, many of the instructions focus on open-ended generation requests (e.g., \textit{``write a poem...''} or \textit{``what should I bring if I were to visit here?''}).

\begin{figure}[h]
    \centering
    \includegraphics[width=0.98\textwidth]{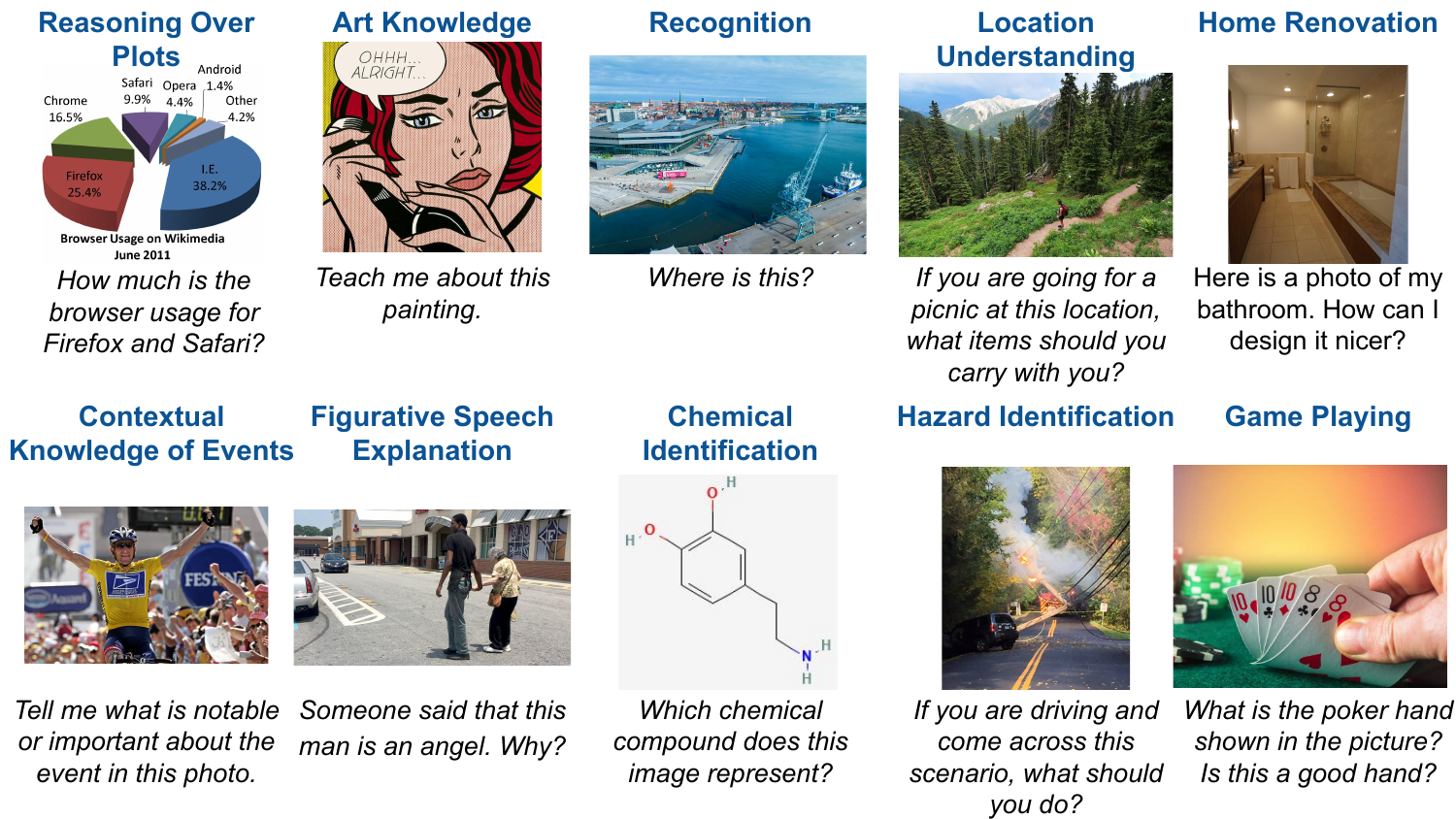}
    \caption{A sample from the 70 instruction families in \benchmarkname{} representing tasks we envision instruction-following vision-language models \textit{should} be able to follow.}
    \label{fig:instruction-families-sample}
\end{figure}

We created \benchmarkname to cover a wide array of ``instruction families''. Our starting point was a set of 70 ``wish-list'' tasks such as ``home renovation'' and ``gardening tips'' collected by the authors:\footnote{We recognize that promising applications may not be covered by our set; and we don't necessarily advocate for deploying models in all cases we cover -- we hope \benchmarkname can help to quantify shortcomings and risks.} each requiring varied high-level skills from recognition to complex reasoning (\Cref{fig:instruction-families-sample}). We derived 25/70 instruction families from benchmark tasks such as Visual Question Answering (VQA) ~\cite{antol2015vqa} and robust change captioning~\cite{park2019robust} into a chatbot-style format (this reformatting differs from prior work~\cite{dai2023instructblip,ye2023mplug,liu2023llava}, as we focus on open-ended chatbot style responses.). Notably, 10 of these repurposed tasks involve multiple images.

We started with 10 images for each instruction family. Our annotators, guided by an example, create a new instruction, and provide a (permissively licensed) image. For each instruction, we next collect \densecaptions{} -- unlike prior work \cite{pont2020connecting, hu2022promptcap} these descriptions are designed not only to describe the image in general, but also, surface information targeted to the instruction. Finally, we use \densecaptions{} to generate a reference candidate output from GPT-4; an additional human verification step discards GPT-4 references deemed to be incorrect.

We conduct a large-scale empirical comparison of multimodal instruction-following models using \benchmarkname (\S\ref{sec:experiments}). We first gather predictions for each instance from 7 candidate models. Then, we collect 5K human judgements of output quality by pitting model outputs head-to-head, and (in a forced-choice setup) crowd-sourcing pairwise preference judgements. This analysis not only reveals significant differences between models (e.g., that LLaVA-13b \cite{liu2023llava} is generally preferred to Panda \cite{su2023pandagpt}), but also, that the human verified references in our corpus are preferred significantly more than the ones generated using multimodal models. We summarize head-to-head comparisons with two metrics: 1) Elo ratings \cite{elo1967proposed,elo_llms}, which provide \textit{relative} ``skill'' rating estimates encoding the probability that model A will be preferred to model B; and 2) win rate versus our references, which provides an \textit{absolute} metric. The best model according to human judgement is LLaMA-Adapter-v2~\cite{gao2023llama}, yet it only wins in a pairwise setting against the reference in 27.4\% of cases.

\begin{figure}[!t]
    \centering
    \includegraphics[width=1.0\textwidth]{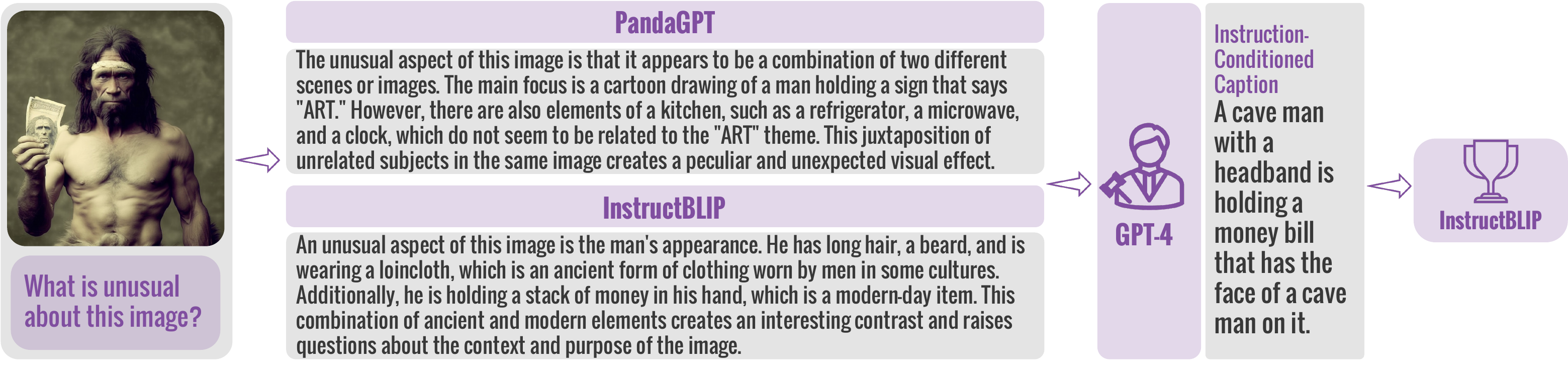}\\
    \caption{ELO-based evaluation for \benchmarkname: Our reference-free approach uses a GPT4 evaluator to compare two instruction-following models with an instruction and a \densecaption. The instance is obtained from an existing dataset, WHOOPS!~\cite{bitton2023breaking}.}
    \label{fig:gpt4_judge}
\end{figure}

Finally, we design an automated evaluation for \benchmarkname, utilizing GPT-4 to rank pairs of model responses based on factors like correctness, relevance, and fluency. Using the \densecaption and the instruction, GPT-4 determines the better response between two options, expediting iteration compared to human preferences. We explore \textit{reference-free} and \textit{reference-backed} versions of this metric.  Compared to various metrics (BLEU-4~\cite{papineni2002bleu}, ROUGE-L~\cite{lin2004rouge}, METEOR~\cite{banerjee2005meteor}, CIDEr~\cite{vedantam2015cider}, and BERTScore \cite{zhang2019bertscore}), our evaluation aligns best with human preferences. For example, it achieves a 94\% agreement rate in the cases where all five annotators agree. See \Cref{fig:gpt4_judge} for a schematic of the process.

While it is difficult to \textit{a priori} envision all possible scenarios under which more performant multimodal chatbots might be used, we hope \benchmarkname can provide a path to improving vision-language models ``in the wild.'' Table \ref{related_work_table} presents a summary of our contributions in comparison to the recent works \cite{xu2022multiinstruct,dai2023instructblip,ye2023mplug,li2023m,xu2023lvlm,liu2023gavie} in the evaluation of multimodal chatbots. We publicly release \benchmarkname data, code, and automatic metrics to facilitate future model evaluations, available in \url{https://visit-bench.github.io/}.

\begin{table}[ht]
\begin{center}
\caption{Comparison with related works for evaluating instruction-following vision-language models. Win-rates* refers to the model win-rates against a reference output/model.}
\label{related_work_table}
\resizebox{\linewidth}{!}{
\begin{tabular}{lccccccc}
\hline
  & MultiInstruct~\cite{xu2022multiinstruct} & Owl~\cite{ye2023mplug}                          & InstructBLIP~\cite{dai2023instructblip}           & M$^3$IT~\cite{li2023m}                                 & LVLM~\cite{xu2023lvlm}                                                                 & GAVIE~\cite{liu2023gavie} & \textbf{VisIT-Bench} \\\hline
Number of Models & 1& 5& 3          & 4 & 8   & 5 & 10            \\
Number of Skills Tested        & 9& 6& 13         & 13 & 47 & 16& 70             \\\hline
Multiple-Images               & \xmark & \cmark               & \xmark         & \xmark& \xmark              & \xmark   & \cmark            \\
 Video  & \xmark & \xmark               & \cmark         & \cmark & \xmark             & \xmark   & \xmark             \\
Multi-Turn Conversations& \cmark & \cmark               & \cmark         & \cmark & \cmark              & \xmark  & \xmark             \\
Multilingual Conversations & \xmark & \cmark               & \xmark         & \cmark & \xmark             & \xmark   & \xmark             \\\hline
Instruction-conditioned Captions                & \xmark & \xmark                & \xmark         & \xmark& \xmark               & \xmark  & \cmark            \\
Chatbot-style Responses& \xmark & \xmark                & \xmark         & \xmark& \xmark                 & \xmark & \cmark           \\\hline
Dataset-specific Evaluation        & \cmark& \cmark               & \cmark        & \cmark & \cmark             & \xmark   & \xmark             \\
Human Evaluation& \xmark & \cmark               & \xmark         & \xmark& \cmark               & \xmark  & \cmark            \\
Auto/GPT-4 Evaluation & \xmark & \cmark                & \xmark         & \cmark & \xmark             &  \cmark     & \cmark            \\\hline
Win-rates* & \xmark & \cmark               & \xmark         & \cmark & \xmark               &  \cmark   & \cmark            \\
\elo Rating     & \xmark & \xmark                & \xmark         & \xmark& \cmark   & \xmark & \cmark           \\\hline
\end{tabular}

}
\end{center}
\end{table}

\section{\benchmarkname: A Real-World Inspired VL Instruction-Following Benchmark}
\label{sec:datasets}
\begin{figure}[h]
    \centering
    \includegraphics[width=\textwidth]{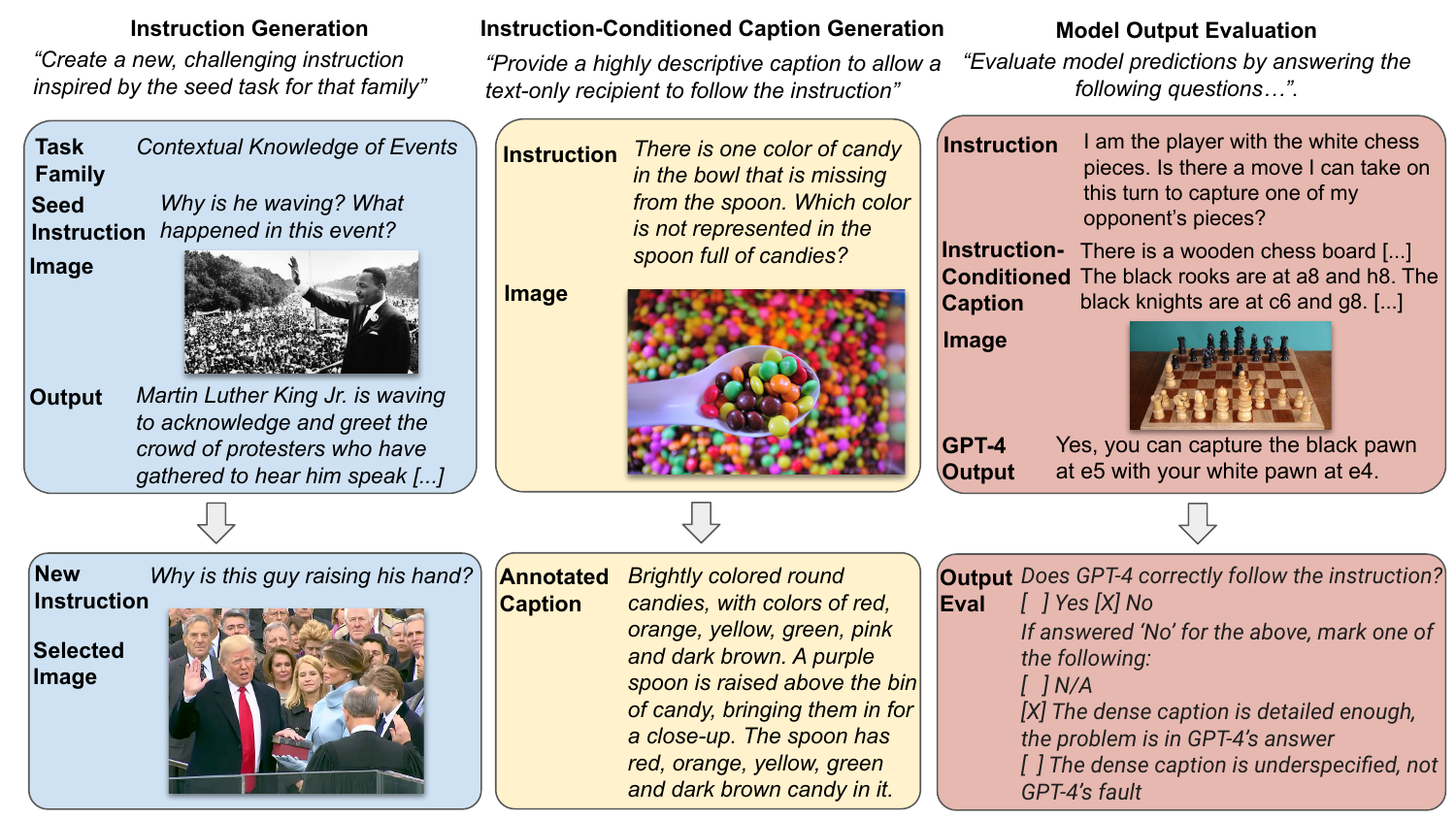}
    \caption{Data collection steps: 
    (1) \textbf{Instruction Generation} - Annotators derive instances from a seed task, see \Cref{fig:data-collection} (left). 
    (2) \textbf{Caption Generation} - Rich \textit{\densecaptions} are produced for GPT-4 references, shown in \Cref{fig:data-collection} (middle). 
    (3) \textbf{Model Evaluation} - GPT-4 responses are human-validated, illustrated in \Cref{fig:data-collection} (right). 
    Top blocks show rater instructions; bottom blocks depict outputs.
    }

    \label{fig:data-collection}
\end{figure}

\benchmarkname was built to emulate real-world applications of multimodal models through image-text tasks, creating an extensive and practical benchmark. These tasks, or `instruction families', are seen as key capabilities of a high-performing vision-and-language model. Although our selections are not exhaustive, they provide a broad basis for evaluating beyond academic benchmarks.
We prioritize family coverage vs. number of instances-per-task. 
The final corpus, comprising 592 instances and 1,159 public images, can be found at \sheetlink{} and \sheetlinkmulti{}. \benchmarkname instances are either from 45 newly assembled instruction families or reformatted from 25 existing datasets (see \Cref{table:existing_datasets}). Notably, 10 instruction families cater to \textit{multi-image} query scenarios (e.g., \Cref{fig:multi_image_example}).


\subsection{Data Collection}
The authors of this work perform an initial annotation step of curating instruction families. For each instruction family not derived from an existing task (45 out of 70), we designate a name for the family (e.g., ``Contextual Knowledge of Events'') and identify an image-instruction pair that exemplifies the category, along with a sample response (``Martin Luther King Jr. is waving to acknowledge and greet the crowd of protesters [...]''). 10 sample familes are in \Cref{fig:instruction-families-sample}. 

The following steps are carried out in collaboration with crowdworkers, who receive an hourly wage of \$18. These steps are outlined in \Cref{fig:data-collection}: (1) taking the image/instruction example as a guiding seed task 
crowdworkers formulate a new instruction that examines the same instruction family (``instruction generation''); (2) crowdworkers create detailed image captions that describe the image and allow an entity, relying solely on this text, to interpret and execute the given instruction successfully (``instruction-conditioned caption generation''); (3) crowdworkers assess the correctness of GPT-4's response to the instruction (``model output evaluation''). We further elaborate on these steps using human annotators below.

\begin{figure}[!t]
    \centering
    \includegraphics[width=\linewidth]{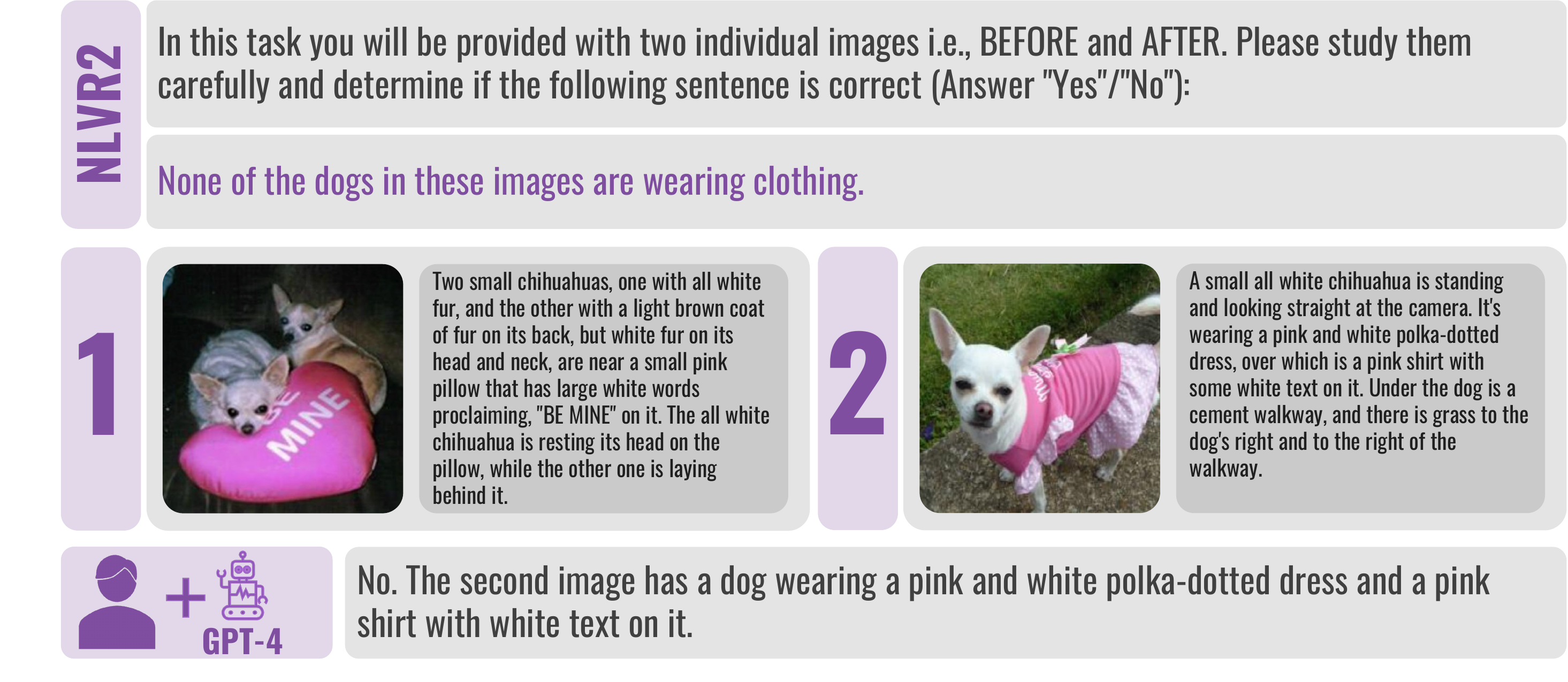}
    \caption{An example of a multi-image instruction task from \benchmarkname{}. Originally sourced from NLVR2~\cite{suhr2018corpus}, this task aims to evaluate visual reasoning skills. The NLVR2 format includes a sentence to analyze, two images, and a binary response. Our adaptation adds a zero-shot instruction prompt, a \densecaption{} for each image and a human-verified GPT-4 response. These enhancements, designed in a chatbot-style interaction, facilitate automatic evaluation of future chatbot responses to this instance.}
    \label{fig:multi_image_example}
\end{figure}

\paragraph{Re-formatting existing datasets.} 25/70 instruction families (corresponding to 25*10=250 instances) are re-formatted versions of existing vision-language tasks (See Appendix~\ref{sec:existing_datasets_list} for full list).\footnote{Users of \benchmarkname should also cite the original datasets.} This process involves re-formatting tasks into chatbot-style instruction/response versions. In re-formatting, we re-write instructions to retain the original task's goal while maintaining the original images, see \Cref{fig:multi_image_example}. These repurposed tasks are integrated into our data collection process, ensuring uniformity between the chatbot-style answers in the full \benchmarkname{} instances and the reinterpreted tasks. 

\paragraph{Instruction Generation.} 
Here, annotators create a new instance from the same instruction family as a given example, along with an instruction and corresponding image. For instance, in \Cref{fig:data-collection} (left), the instruction family is ``Contextual Knowledge of Events'', and the example instruction is \textit{``Why is he waving? What happened in this event?''} alongside an image of Martin Luther King, Jr. 
To collect images, annotators were instructed to use Openverse (https://openverse.org/) for Creative Commons licened images.

\paragraph{Instruction-Conditioned Caption Generation.} 
Annotators are provided with the image and instruction, and are tasked to construct a caption that is rich enough to allow an entity, solely receiving the text they author, to follow the instruction. This caption will later facilitate GPT-4 reference candidate generation, and will be used for text-only auto-evaluation. We call these instructions \textit{\densecaptions}. See \Cref{fig:data-collection} (middle) for an example: an annotator doesn't just mention the skittles and a spoon, but, given the query regarding specific colors, they indicate the exact colors in detail.

\paragraph{Model Output Evaluation.} The goal of this stage is to gather human-validated reference chatbot responses for each multimodal instruction query. We initially obtain response candidates from GPT-4 given the instruction and the \densecaption. GPT4's prompt is: \textit{``Consider an image depicted by: <caption>'. Now, briefly follow this instruction, and you can add a short explanation: <instruction>'. Response: } This prompt is employed for both single and multiple image instances, with appropriate modifications for the latter. Then we verify each response with human annotators.\footnote{An alternate annotation scheme would have been to task annotators to write target responses from scratch. The rationale for using GPT-4 verification instead is derived from prior results that show promising human-machine collaboration of this form \cite{wiegreffe2021reframing}.} If a response is marked incorrect, the annotator identifies whether the issue lies with the detail level of the \densecaptions or with GPT-4's response itself. For \benchmarkname, we discard any case marked as incorrect for either reason.\footnote{The annotators are also tasked to screen for any offensive, unsound, or harmful advice present in the responses. We did not find or discard any instances.}  An example is given in \Cref{fig:data-collection} (right), where GPT-4's candidate reference response aims to answer a question about a chess position (which it does so incorrectly, and thus, the instance is discarded).

\subsection{Data Collection Annotation and Results}
\label{sec:sec_with_caption_and_reference_correctness}
We conduct the data collection steps in \Cref{fig:data-collection} using Amazon's Mechanical Turk (MTurk) platform. Prior to annotating, each MTurk worker passed a qualification test, which involved five to ten sample tasks designed to assess their ability to generate high-quality annotations. 
More detailed information about the execution process and full user interface examples can be found in Appendix~\ref{app:template}.

\begin{wraptable}{R}{0.62\textwidth}
\centering
\caption{Human rating metrics for the \benchmarkname{} dataset: overall, single-, and multi-image tasks.}
\label{tab:annotation_metrics}
\begin{tabular}{lrrr}
\toprule
\textbf{Metrics} & \textbf{Overall} & \textbf{Single} & \textbf{Multi} \\
\midrule
GPT-4 Correct (\%) & 87.3 & 91.5 & 63.0 \\
Problem in Caption (\%) & 4.0 & 3.6 & 6.0 \\
Problem in GPT-4 (\%) & 7.7 & 3.8 & 30.0 \\
\bottomrule
\end{tabular}
\end{wraptable}

Our annotation results are summarized in \Cref{tab:annotation_metrics}. We measure the throughput of the collection and filtration pipeline. For single-image instances, our pipeline's yield was 91.5\% from the original candidate set. However, the success rate dropped to 63.0\% in the more complex multi-image tasks, accompanied by an uptick in issues either in the captions (6.0\%) or GPT-4's responses (30.0\%). This drop suggests that multi-image queries may pose a more difficult data collection challenge.

\section{\benchmarkname Analysis}
\label{sec:analysis}
We analyze the tasks, images, and \densecaptions of \benchmarkname.

\subsection{Are \densecaptions necessary?}
\begin{figure}[!t]
    \centering
    \includegraphics[width=\textwidth]{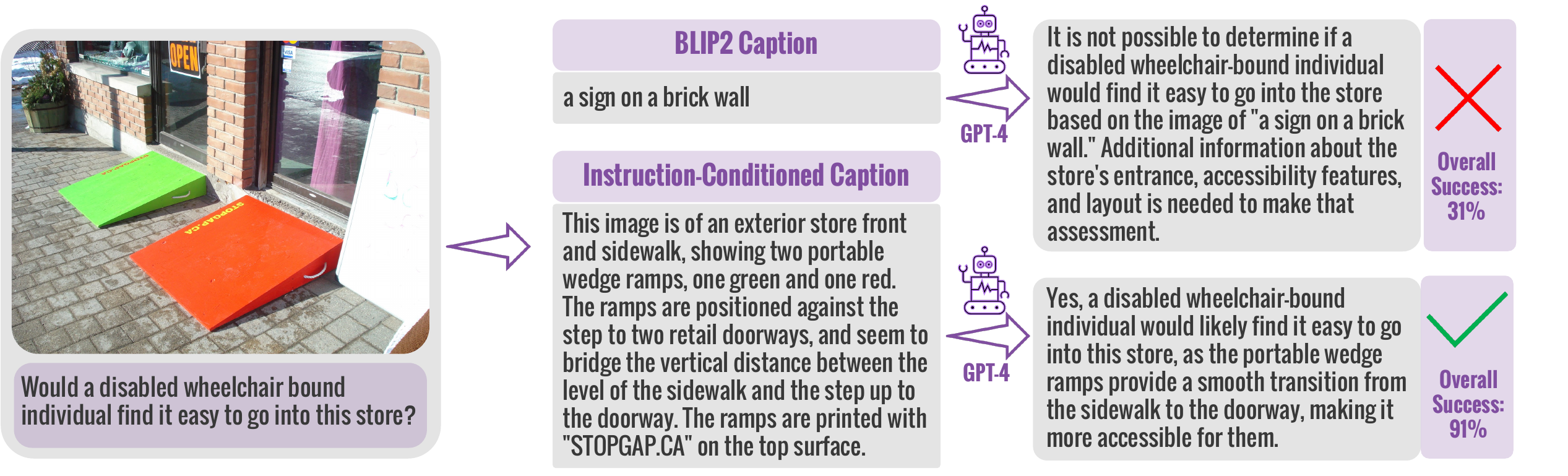}\\
    \caption{This experiment evaluates the value of \densecaptions in accurate instruction-following tasks. Given an image and instruction, GPT-4 generates responses using both a \densecaption and a less detailed BLIP-2 \cite{li2023blip} caption. The latter's imprecision leads to an error, emphasizing the need for detailed, task-specific captions. }
    \label{fig:fig_blip2_ablation}
\end{figure}
To clarify the role of the \densecaptions we collect, we conducted an experiment covering 150 single-image instances. Instead of using our \densecaptions, we use BLIP2~\cite{li2023blip} image captions, which is a state-of-the-art image captioning model. We extract image captions, and feed them to GPT-4 as detailed earlier, to provide a text-based chatbot response. This process is depicted in \Cref{fig:fig_blip2_ablation}.

We manually evaluated whether the resulting output accurately followed the instructions. We find that while \densecaptions led to correct outputs in 91\% of the cases, the success rate fell to 31\% when using BLIP2 captions (\Cref{tab:annotation_metrics}). These results highlight the importance of \densecaptions in the construction of \benchmarkname, and show that the instances in our dataset are sophisticated enough such that most are not solvable by using a simple Socratic model \cite{zeng2022socratic} baseline of caption $\rightarrow$ LLM. 

\subsection{What skills are required for \benchmarkname?}
\begin{wrapfigure}{r}{0.58\textwidth}
    \centering
\includegraphics[width=0.58\textwidth]{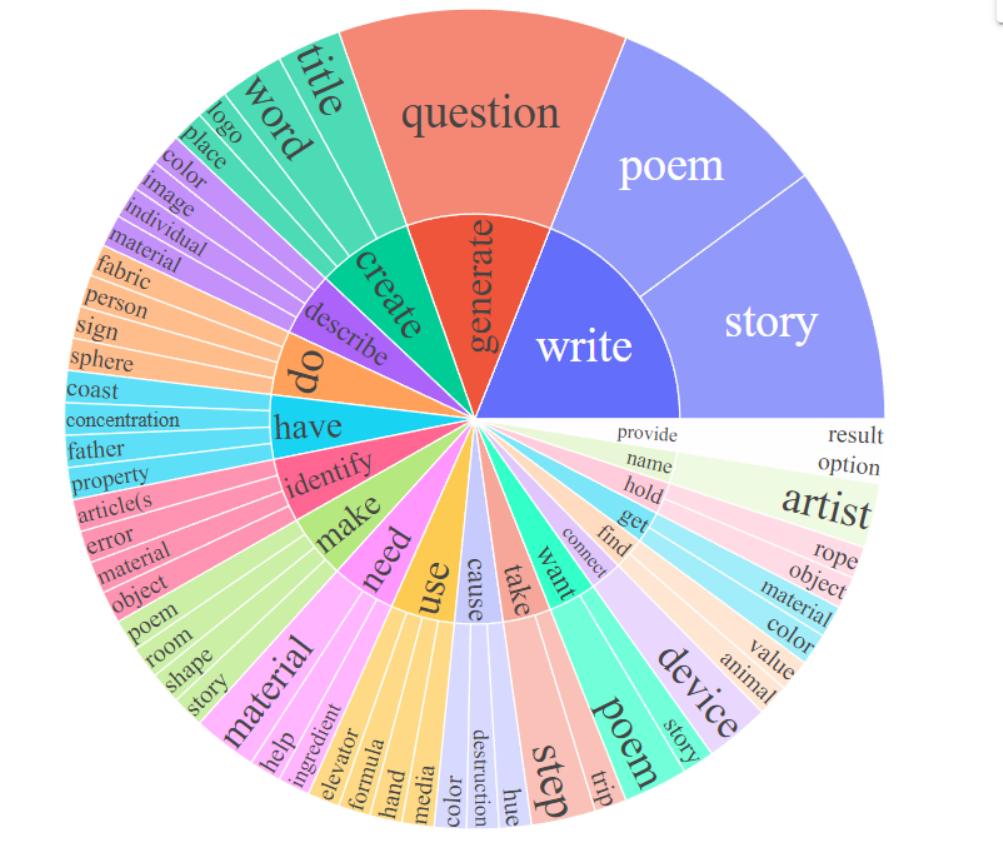}
    \caption{Most frequently occurring verbs (inner circle) and their top 4 direct nouns (outer circle) in the \benchmarkname instructions.}
    \label{fig:wheel-instructions}
\end{wrapfigure}

The full list of instruction families we cover are in Appendix \Cref{tab:skills}.
Following \cite{wang2022self}, for the \benchmarkname instructions, we extract the most frequent root verbs and their direct nouns (a full plot is in \Cref{fig:wheel-instructions}). The most common include: \textit{`answer question'}, \textit{`write story/poem'}, \textit{`create title'}, etc. There's also a long-tail of diverse requests that demand comprehension, commonsense, and cross-modal understanding, e.g., \textit{`identifying objects'} to \textit{`need ingredient'} to \textit{`connect device'}. Additional qualitative examination reveals a range of underlying skills required ranging from \textit{`emotion identification'} to complex reasoning tasks such as \textit{`paper folding'}. 

\subsection{What is contained in \benchmarkname images?}
We detect all the COCO \cite{lin2014microsoft} objects present in the images from our dataset using Yolov5-L \cite{redmon2016you};
The most common detected objects in \benchmarkname are ``person'' ({\raise.17ex\hbox{$\scriptstyle\sim$}} 900 detections), chair, and car ({\raise.17ex\hbox{$\scriptstyle\sim$}} 100). But, a long tail of rarer objects exists as well: full distribution in Appendix \Cref{fig:objects-visit}. Overall, to perform well at \benchmarkname, a model must account for a broad range of scenes and objects.


\section{Experiments}
\label{sec:experiments}

We evaluate a range of state-of-the-art publicly accessible vision-and-language chatbots on the 592 instances in \benchmarkname. In \S \ref{setup}, we provide the details of the instruction-following models in our benchmark. Following this, we collect the human preferences for pairwise model generations to achieve a human-guided \elo ranking and the win-rates against the reference of the models in \S \ref{human_guided}. We then develop automatic evaluation on \benchmarkname in \S \ref{auto_eval}, that can be scaled and improved given new and improved models. Finally, we establish the trustworthiness of our automatic evaluation method by performing agreement analysis with the human judgments in \S \ref{agreement}

\subsection{Models}
\label{setup}
We evaluate LLaVA-13B~\cite{liu2023llava}, InstructBLIP-13B~\cite{dai2023instructblip}, MiniGPT4-7B~\cite{zhu2023minigpt}, mPLUG-Owl-7B~\cite{ye2023mplug}, LlamaAdapter-v2-7B~\cite{gao2023llama}, PandaGPT-13B~\cite{su2023pandagpt}, VisualChatGPT~\cite{wu2023visual}, Multimodal GPT~\cite{gong2023multimodal}, OpenFlamingo v1~\cite{anas_awadalla_2023_7733589,awadalla2023openflamingo} , Otter v1~\cite{li2023otter}, Lynx~\cite{huang2023sparkles} and idefics~\cite{laurencon2023obelics}. For the execution-based  VisualChatGPT~\cite{wu2023visual}, we implement a chat window for each sample, hold inputs and intermediate chains of thoughts and actions in memory, and feed the images and the instruction sequentially. For OpenFlamingo~\cite{anas_awadalla_2023_7733589} and Otter~\cite{li2023otter}, we feed the image(s) and the instruction in an interleaved format. For the others, we feed the image to the vision feature extractor and feed the instruction as a prompt to the text encoder.\footnote{Following the authors' instructions, we run all models using default settings to obtain the best possible responses. We include specific samples for reproducibility. We acknowledge hyperparameter impact and are willing to reassess submissions to \benchmarkname if conditions were sub-optimal.}

\subsection{Human Evaluation}
\label{human_guided}
\label{sec:sec_with_human_evaluation}

We collect 5K pairwise human preference judgements across an initial set of 6 models and the human-verified references. For 1K uniformly randomly sampled tuples of (query, model A, model B), we collect 5 crowdworker judgements each. Preferences are collected in a ``forced choice'' setting, annotators are instructed to decide based on accuracy, helpfulness, and detail. We provide the template for the human annotation process in Appendix \Cref{fig:templates-model-rating}.  We summarize the results with two metrics: 

\textbf{Relative metric: \elo} We follow \cite{elo_llms} and compute \elo ratings, treating each pairwise human judgement as a ``match.''\footnote{We use the following code/hyperparameters for \elo ratings: \url{https://github.com/lm-sys/FastChat/blob/main/fastchat/serve/monitor/elo_analysis.py}} The difference between the \elo ratings of two different models provides an estimate for the win probability when pitting model A vs. model B. More details are in Appendix \ref{app:elo_rating}. 

\textbf{Absolute metric: Win rate vs. reference.} We provide a win-rate vs. the human-verified reference. We use the 1.4K pairwise human judgments where one of A or B is the reference. We report the percent of cases where the human judge prefers the output from that model vs. the human-verified GPT-4 reference output. Because we do not allow for ties in our forced-choice setup, if the annotator believes the responses are of equal quaity, they choose one arbitrarily. 

\begin{table}[!t]
\centering
\caption{Human scoring results for the models,
shown as both an ELO rating and win-rate against the reference. In total, this summarizes 5.0K pairwise human judgments. matches column indicates the number of total matches in which a particular model participates. Win-rate vs. reference indicates the win-rate of a model against the reference outputs.}
\label{tab:table_human_scoring_results}
\resizebox{1\textwidth}{!}{%
\begin{tabular}{l l r r c}
\toprule
\textbf{} & \textbf{Model} & \textbf{\elo}& \textbf{matches} & \textbf{Win-rate vs. reference} (w/ \# ratings) \\
\midrule
Single Image & Human Verified GPT-4 Reference & 1223 & 1439& -- \\ 
& LLaVA (13B)  & 1085 & 1462& 26.23\% (n=244) \\
& LlamaAdapter-v2 (7B) & 1061 & 1507& \textbf{27.41\%} (n=259) \\
& mPLUG-Owl (7B) &  995& 1345 & 14.95\% (n=214) \\
& InstructBLIP (13B) & 957 & 1315 & 12.37\% (n=194) \\
& MiniGPT-4 (7B) & 893& 1513 & 14.72\% (n=299) \\
& PandaGPT (13B) & 786 & 1441& 10.48\% (n=229) \\
\midrule
Multiple Images & Human Verified GPT-4 Reference & 1193 & 210 & -- \\
& mPLUG-Owl & 997 & 190 & 15.38\% (n=78) \\
& Otter v1 & 917 & 147 &  3.17\% (n=63) \\
& OpenFlamingo v1& 893 & 171 & 4.35\% (n=69)\\
\bottomrule
\end{tabular}
}
\end{table}

\paragraph{Results} 
Table~\ref{tab:table_human_scoring_results} contains the \elo and win-rate vs. reference. In terms of \elo, the Human Verified GPT-4 reference achieves a higher rating than all alternatives, validating the quality of our reference set: concretely, for our \elo settings, the reference (\elo=1223) has an estimated win-rate over one of the best performing models, LLaVA, (\elo=1085) of 69\%, and an estimated win rate of 93\% against the lowest performing model in this setup, PandaGPT (\elo=786). This result can partly be explained by the training process of the underlying models:  The improved performance of LLaVA (13B) might be attributed to its fine-tuning process, which utilized 150K instruction-tuning data that is rich in both diversity and quality.
Interestingly, despite achieving a slightly lower \elo (the computation of which is based on \textit{all} head-to-head ``matches'', rather than just ones against the human reference), LlamaAdapter-v2 (7B) wins with the highest rate against the reference. However, the complexity and variety of models and tasks in \benchmarkname{} makes it challenging to definitively pinpoint the factors influencing performance. While we make a preliminary attempt to unravel these intricacies in \Cref{auto_eval}, a comprehensive understanding will necessitate more nuanced and extensive future research.

\subsection{Automatic Evaluation and Leaderboard}
\label{agreement}
\label{auto_eval}

\begin{table}
\centering
\caption{Current reference-free Elo rankings as of September 25th, 2023. In total, these rankings summarize 31,735 ``matches'' between models; each match consists of 2 queries to GPT-4. Because \benchmarkname is dynamic, these rankings are updated as more models are added to the leaderboard, and more pairs of models are evaluated head-to-head for more instances.}
\label{tab:table_updated_auto_scoring_results}
\resizebox{\textwidth}{!}{%
\begin{tabular}{l l r r c}
\toprule
\textbf{Category} & \textbf{Model} & \textbf{Elo} & \textbf{\# Matches} & \textbf{Win vs. Reference}  (w/ \# ratings) \\
\midrule
\multirow{15}{*}{\centering Single Image} 
& Human Verified GPT-4 Reference  & 1,382 & 5,880 & --- \\
& LLaVA-Plus (13B) & 1,203 & 678 & 35.07\% (n=134) \\
& LLaVA (13B) & 1,095 & 5,420 & 18.53\% (n=475) \\
& mPLUG-Owl (7B) & 1,087 & 5,440 & 15.83\% (n=480) \\
& LlamaAdapter-v2 (7B) & 1,066 & 5,469 & 14.14\% (n=488) \\
& Lynx(8B) & 1,037 & 787 & 11.43\% (n=140) \\
& idefics (9B) & 1,020 & 794 & 9.72\% (n=144) \\
& InstructBLIP (13B) & 1,000 & 5,469 & 14.12\% (n=503) \\
& Otter v1 (9B) & 962 & 5,443 & 7.01\% (n=499) \\
& VisualGPT (Da Vinci 003) & 941 & 5,437 & 1.57\% (n=510) \\
& MiniGPT-4 (7B) & 926 & 5,448 & 3.36\% (n=506) \\
& Octopus V2 (9B) & 925 & 790 & 8.90\% (n=146) \\
& OpenFlamingo V1 (9B) & 851 & 5,479 & 2.95\% (n=509) \\
& PandaGPT (13B) & 775 & 5,465 & 2.70\% (n=519) \\
& Multimodal GPT & 731 & 5,471 & 0.19\% (n=527) \\
\midrule
\multirow{4}{*}{\centering Multiple Images} 
& Human Verified GPT-4 Reference & 1,192 & 180 & - \\
& mPLUG-Owl & 995 & 180 & 6.67\% (n=60) \\
& Otter v1 & 911 & 180 & 1.69\% (n=59) \\
& OpenFlamingo v1 & 902 & 180 & 1.67\% (n=60)\\ 
\bottomrule
\end{tabular}%
}
\end{table}

\begin{figure}[!b]
    \centering
    \includegraphics[width=\textwidth]{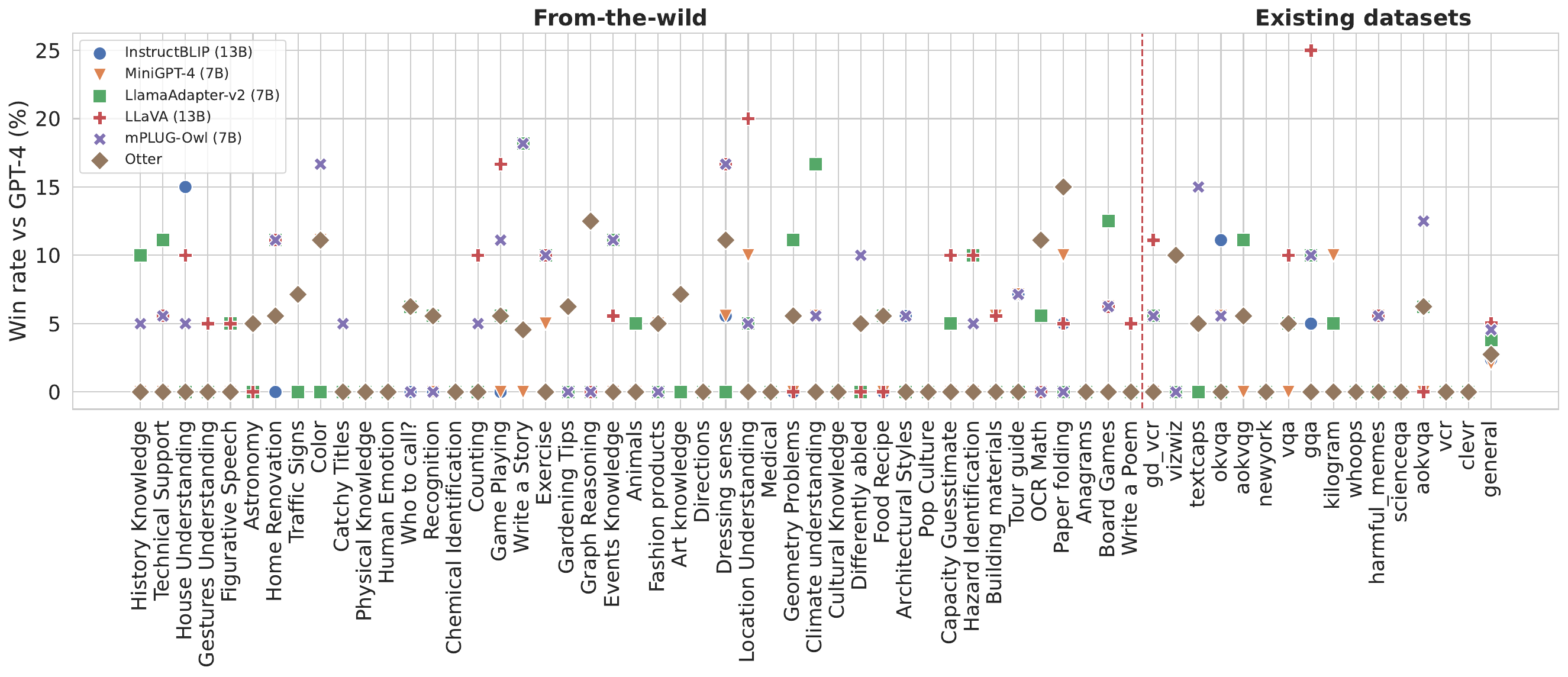}\\
    \caption{Reference-free assesment win rate vs. human-verified GPT4 response for each instruction category. Axes: win rate (Y), instruction categories (X). Categories are from-the-wild or existing datasets. \benchmarkname{} facilitates analysis of diverse instruction tuning tasks. }
        \label{fig:fig_winrate_per_category}
\end{figure}

\begin{wrapfigure}{r}{0.5\textwidth}
    \centering
    \includegraphics[width=0.48\textwidth]{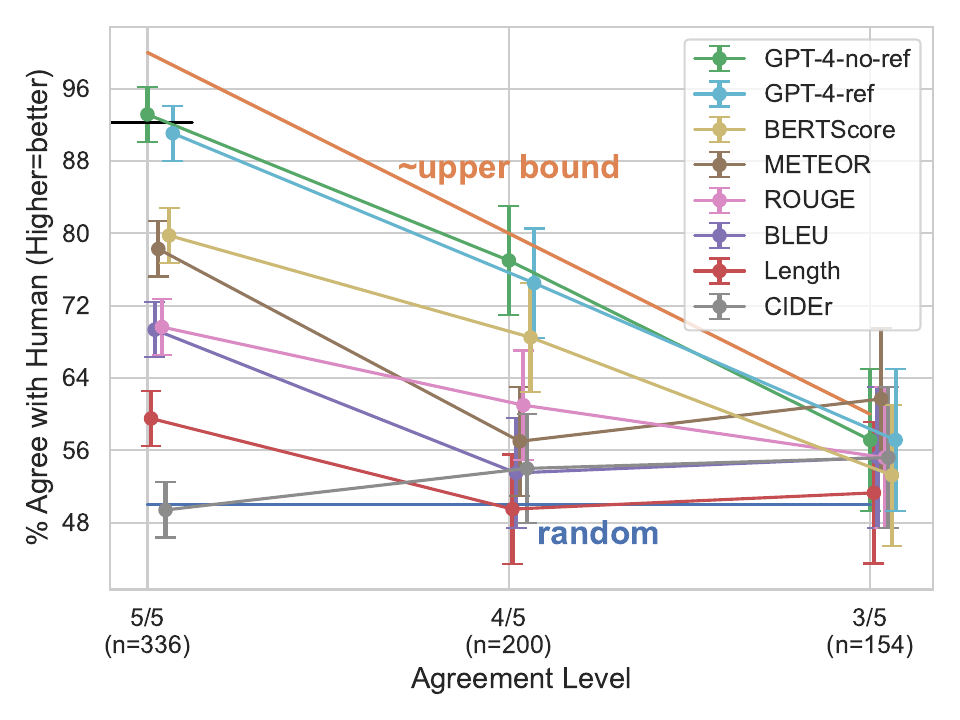}
    \caption{Correlations between evaluation metrics and human preferences are ranked in performance order, with our reference free evaluation (GPT-4-no-ref) showing the strongest alignment. Bottom line: random chance (50\%), top line: upper performance bound.}
    \label{fig:metrics_correlations}
\end{wrapfigure}

Because it is costly to gather human pairwise preference judgements for new model submissions, to support faster model development, we seek an automatic evaluation procedure that produces high correlation with our human evaluation setup.

\paragraph{Automatic evaluation metric candidates.} We consider several existing reference-backed evaluation metrics:
BLEU-4~\cite{papineni2002bleu} , ROUGE-L~\cite{lin2004rouge}, METEOR~\cite{banerjee2005meteor}, CIDEr~\cite{vedantam2015cider}, and BERTScore \cite{zhang2019bertscore}, we use the RoBERTa-Large english version \cite{liu2019roberta}, treating the human-verified GPT-4 reference as the evaluation reference. We additionally report two baseline metrics: random, which assigns a random score without accounting for the candidate, and length, which assigns a score equal to the number of non-whitespace tokens in the candidate. Beyond existing metrics and baselines, following the recent line of work utilizing API-accessed LLMs with a prompt for automatic evaluation \cite{vicuna2023,dubois2023alpacafarm},
we consider two GPT-4
\cite{gpt4} backed evaluation metrics. 

Specifically, we provide the LLM with:
1) a system prompt describing the desired evaluation behavior;
2) the \densecaption for the image;
3) the instruction to be followed;
and 4) two candidate generations dubbed ``Response A'' and ``Response B''. We also consider a reference-backed version where the human-verified reference is provided as well.
We provide our prompts in \Cref{app:prompts}.
To mitigate potential biases in ``A'' and ``B'' positioning, for all pairs of candidates, we run two queries covering both possible orderings. Our prompt encourages the model to think step-by-step so that its chain-of-thought process is made explicit \cite{wei2022chain,kojima2022large}. Despite strongly encouraging the model to select between the two references in a forced-choice setup, it sometimes refuses and outputs ``tie'' which we account for later. We call the reference-free version of this metric ``GPT4-no-ref'', and the reference-backed version of this metric ``GPT4-ref''. 
\paragraph{Evaluating evaluation metrics.} We measure the correlation between the candidate metrics and human judgements using a pairwise framework. Specifically, we use a subset of the 5K pairwise human judgements in \S~\ref{sec:sec_with_human_evaluation}. For 690 pairwise instances where both candidate instances are model-generated (rather than human-verified references), we have 5 pairwise judgements from crowd-workers. For 336 pairs, there is 5/5 agreement, for 200 pairs, there is 4/5 agreement, and for 154 pairs, there is 3/5 agreement. For each metric, we measure the percent of time the metric is able to accurately reconstruct a majority vote judgement from the 5 crowdworkers. The newly proposed GPT-4 based metrics sometimes outputs ``tie'' (this happens in ~10-15\% of cases overall) -- for fair comparison with the other metrics in forced choice setting, we randomly choose one of the two options when GPT-4 reports a tie. 

The results are in \Cref{fig:metrics_correlations}, with GPT-4-no-ref best aligns with human correlation.
The best performing metric is our newly proposed GPT-4 based metric, which accurately reconstructs majority-vote pairwise human judgments better than alternatives ($p<.05$; binomial proportion CI nonoverlapping). For example, for instances where 5/5 annotators agree, GPT4-no-ref, with no reference, accurately reconstructs human judgment 93\% of the time, whereas the next best metrics BERTScore/METEOR/ROUGE-L reconstruct accurately 80\%/78\%/70\% of the time; among the metrics we consider, these are reasonable options for static/offline evaluation without relying on OpenAI API access, especially when compared to our length baseline metric, which achieves only 60\%. Notably, the reference-backed version of the newly proposed GPT-4 based metric achieves comparable (but slightly worse) performance compared to the reference-free version. Thus, we adopt the reference-free version, which additionally enables us to place the references themselves into the the \elo setup, because they are not used in the prompts.

\textbf{System-level Correlation.} We summarize the LLM's pairwise judgements using the same metrics as introduced in \S\ref{sec:sec_with_human_evaluation}, \elo ratings and win rate vs. reference, but instead of using a human judge, we use our reference-free GPT-4 based metric. The results are in \Cref{tab:table_auto_scoring_results}. Notably, among the 7 systems for which we gathered human ratings for, the automatic metric produces the same ordering compared to human evaluation ($\rho=1.0$, $p<.01$).

\textbf{Shortcomings of proposed metric.} While the relative ranking of models produced by the automatic metric correlates strongly with the ranking produced by human judgements, the win rate vs. reference according to human judgement (\Cref{tab:table_human_scoring_results}) are higher overall compared to the win-rate vs. reference according to the automatic metric \Cref{tab:table_auto_scoring_results}. One plausible explanation for this discrepancy is that GPT-4, as an evaluation model, may prefer responses that closely match its own response distribution. 

\textbf{Per-category results.} In \Cref{fig:fig_winrate_per_category}, we plot the win-rate vs reference for the models across all the single-image instruction families. We find that there is no model that performs the best and worst across all the instruction families. Thus, \benchmarkname aids in highlighting the strengths and weaknesses of the instruction-following models along various real-world use-cases.




\section{Related Work}
\label{sec:related}

\textbf{Multimodal Models for Image-Text Understanding:} Recently, the field of machine learning has experienced a rapid proliferation of new models which can perform various image-text tasks \cite{Alayrac2022FlamingoAV, li2023blip, liu2023llava, pandey2022cross, su2023pandagpt, dai2023instructblip}. This growth has been driven by several factors, including the emergence of large-scale multimodal datasets (e.g. LAION-5B \cite{schuhmann2022laion}, Multimodal C4 \cite{anas_awadalla_2023_7733589}), improved software and hardware frameworks, and advances in modality-specific models such as language models (e.g., \cite{touvron2023llama}). 
Our work specifically evaluates models which can generate textual outputs, given one or more images, and text. Recent examples of such models include LLaVA \cite{liu2023llava}, mPLUG-Owl \cite{ye2023mplug}, InstructBLIP, LLaMA-Adapter, Flamingo \cite{Alayrac2022FlamingoAV} and OpenFlamingo \cite{anas_awadalla_2023_7733589}, PandaGPT \cite{su2023pandagpt}, and GPT-4 \cite{gpt4} (which reports multimodal capabilities but has not yet seen a release of the multimodal variant).

\textbf{Instruction Following:} ``Instruction-following'' is an emerging paradigm for training models via language, where instead of being trained to complete only a single, fixed task (such as image classification or captioning), models are trained to follow textual instructions that describe an arbitrary task, with the aim of generalizing to novel instructions. Examples of instruction-following models include Alpaca \cite{alpaca}, LLaMA-Adapter \cite{gao2023llama}, Koala \cite{koala_blogpost_2023}, InstructBLIP \cite{dai2023instructblip}, LLaVA \cite{liu2023llava}, and mPLUG-owl \cite{ye2023mplug}. As the downstream capabilities of these models are influenced by the quality of the training dataset, there has also been extensive work on developing instruction-following datasets \cite{wang2022self, peng2023instruction, yin2023dynosaur, liu2023llava, zhou2023lima} .

To build these models, two broad approaches have been shown to be effective. One approach focuses on leveraging existing pretrained task-specific tools such as image captioners \cite{li2023blip}, object detectors \cite{li2022grounded} and text-to-image generators \cite{rombach2022high} by either creating multimodal prompt interfaces \cite{wu2023visual,yang2023mm} or by executing LLM-generated programs \cite{suris2023vipergpt,gupta2023visual,lu2023chameleon}. The other approach \cite{liu2023llava,gao2023llama,gong2023multimodalgpt,li2023otter,huang2023language,ye2023mplug,anas_awadalla_2023_7733589} focuses on building a single pretrained model that can follow instructions by supervised finetuning on multimodal vision-language data. 

Despite the success of both these approaches on the existing vision-language datasets e.g., VQA, GQA, Image Captioning \cite{antol2015vqa,hudson2019gqa,lin2014microsoft}, there is a lack of a high-quality benchmarking dataset for multimodal instruction-following tasks that reliably replicates the way in which humans would interact with multimodal chatbots in the wild. Similar to the image-text models discussed above, many instruction-following models have been released directly as open-source without undergoing peer review or thorough evaluation. As a result, the effectiveness of these models for many tasks is not well-understood.

\textbf{Benchmarks for Machine Learning:} High-quality evaluation datasets have served both to (re)assess, and to accelerate, progress on many machine learning tasks \cite{liao2021we}. For example, our work draws particularly from the fields of computer vision and natural language processing, where benchmarking datasets have been critical drivers of progress. On the vision side, datasets such as ImageNet \cite{russakovsky2015imagenet} and CIFAR \cite{krizhevsky2009learning} have proven to be critical yardsticks of progress. On the language side, benchmarks such as SQuAD \cite{rajpurkar2018know}, SST \cite{socher2013recursive}, GLUE/SuperGLUE \cite{wang2018glue, wang2019superglue} and more \cite{eval-harness, liang2022holistic} seen wide use. Recent work has indicated that improvements on these high-quality benchmark datasets is \emph{not} the result of overfitting, and is a reliable indicator of genuine progress beyond the benchmark data \cite{miller2020effect, recht2018cifar, recht2019imagenet, roelofs2019meta}. 

However, high-quality benchmarking datasets and evaluation methods do not yet exist for multimodal instruction-following. 
As a result, it is difficult to assess progress in this direction, which both reduces the field's ability to identify true breakthroughs and increases vulnerability to potential pitfalls of evaluation that have hampered progress in other areas of machine learning \cite{liao2021we,yin2023survey}.

\section{Conclusion}
\label{sec:conclusion}
We introduce \benchmarkname, a dynamic benchmark providing a broad 
evaluation of multimodal chatbots' capabilities. Going beyond prior efforts, \benchmarkname's collection process centers potential real-world use cases, and 70 diverse instruction families encompassing a range of tasks from recognition to complex reasoning. Our benchmark not only offers human-verified reference outputs for all examples but also gives an \elo-based ranking system for multimodal chatbots that correlates with human judgements. Our experiments reveal a gap between model and human performance.
We release data, code, and automatic metrics, encouraging community involvement.
We hope \benchmarkname can provide a new quantification of progress and shortcomings of multimodal AI systems. 

\section{Limitations}
\label{sec:limitations}

Although \benchmarkname covers a wide spectrum of potential use-cases, it does not incorporate every possible vision-language task. We hope to add more categories of tasks over time. In terms of dialogue, \benchmarkname concentrates on single-turn instances with one instruction and response. This does not encompass multi-turn interactions between users and chatbots, which presents a promising direction for future research. Our study focuses on image-text modalities. Future extensions could expand the scope to include other modalities like audio and video, enabling a more comprehensive evaluation. Additionally, while the dataset offers a wide variety of tasks, a larger number of examples per category could provide more depth. Finally, while our GPT-4 based metric correlates well with human judgement both at the instance level and at the system level, we see some evidence that the GPT-4 based metric has a stronger preference for GPT-4 based generations compared to humans. Thus, models which train, e.g., by distilling from GPT-4 outputs, may have an unfair advantage on our evaluation.

\section*{Acknowledgements}
We thank Pang Wei Koh, Ashima Suvarna, Nitzan Guetta and Roee Aharoni for their valuable feedback. Hritik Bansal is supported in part by AFOSR MURI grant FA9550-22-1-0380.
RT is supported by the NSF GRFP under Grant No. DGE 1656518.
\bibliographystyle{unsrtnat}
\bibliography{references}

\appendix

\vfill\null
\clearpage
\appendix
\section*{Appendix}
\label{sec:appendix}
\section{License and Intended Use}
The \benchmarkname dataset, along with its various contributions such as instructions, reference outputs, and model ranking annotations, is licensed under the Creative Commons Attribution 4.0 International License (CC BY 4.0). This license applies to all the images we have directly contributed, each of which carries a public license specification in the ``public images metadata'' field within the dataset sheets. However, the dataset also incorporates images sourced from pre-existing collections. For these images, the original licensing terms are respected and remain applicable.

\benchmarkname's primary purpose is to function as a dynamic benchmark that continuously evolves and evaluates instruction-following vision-language models. In the current landscape, commercial chatbots are often trained on non-disclosed and non-public datasets, which raises concerns about potential data contamination and inadvertent training on our evaluation data~\cite{brown2020language}. This risk is further highlighted by recent studies~\cite{magar2022data,jacovi2023stop}. To mitigate such concerns, we have chosen to withhold the complete \benchmarkname test set from public disclosure, while still making the images and instructions available for direct download. Researchers, however, can utilize \benchmarkname to its full potential as a dynamic benchmark by submitting their model predictions for evaluation. We will assess their models using the undisclosed test set, ensuring the ongoing evolution of the benchmark. Moreover, we are open to releasing the test data upon receiving reasonable and justified requests, particularly when additional analysis is necessary, provided that requesters agree to our non-contamination policy which prohibits the use of this data for training commercial chatbots. This approach strikes a balance between the need for robust model evaluation and the mitigation of potential data contamination. 

\appendix
\section{Dataset Analysis}
\begin{figure}[h]
    \centering
    \includegraphics[scale=0.45]{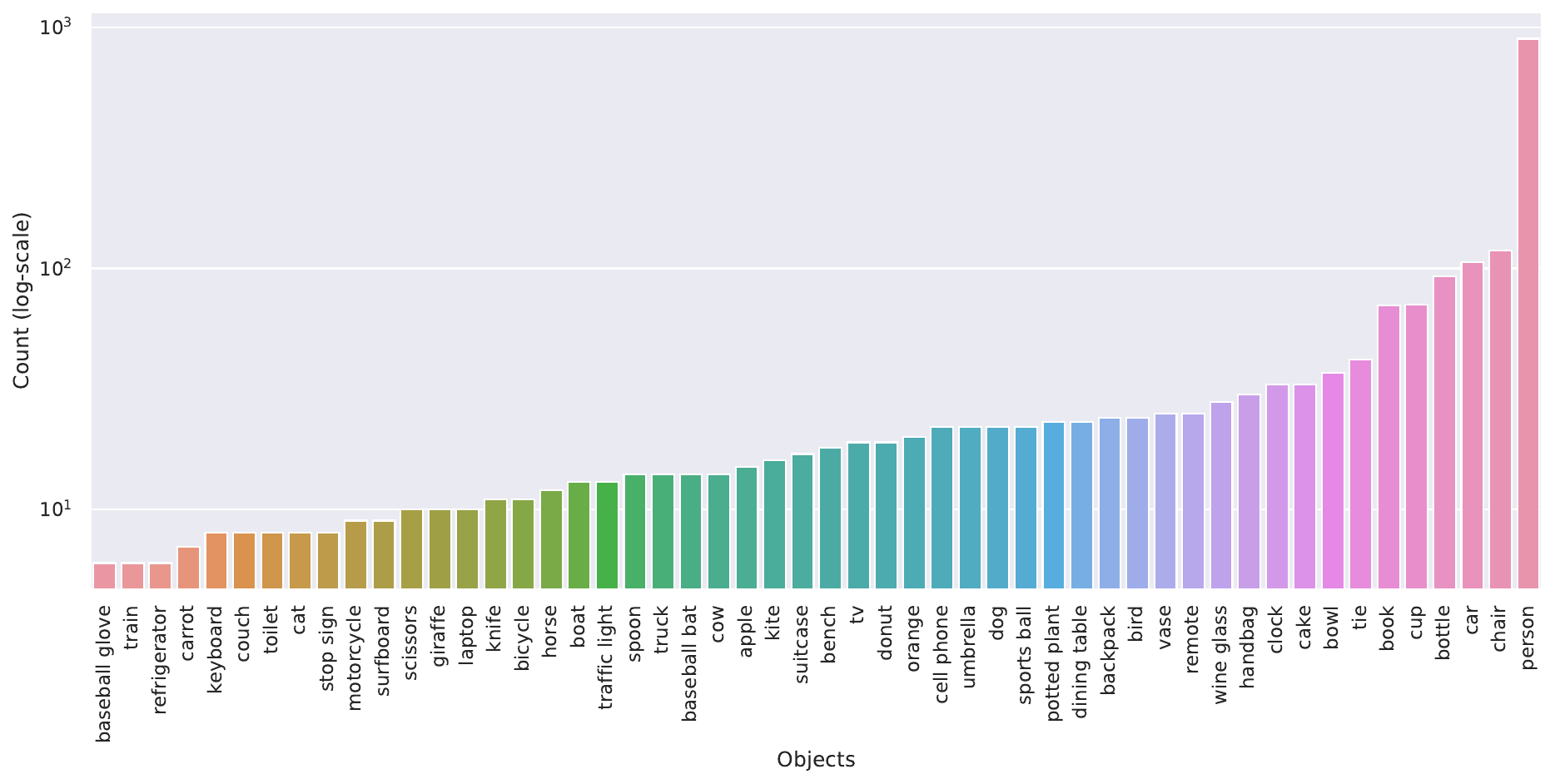}
    \caption{Count of the various COCO objects present in the \benchmarkname images on a log-scale. The object detection was performed by Yolov5-Large \cite{redmon2016you}. We observe that the object `person' occurs most of the time in the dataset images i.e., 875 times.}
    \label{fig:objects-visit}
\end{figure}


\section{Interfaces for Collecting Human Annotations}\label{app:template}
\begin{figure}[ht]
    \centering
    \includegraphics[width=\linewidth]{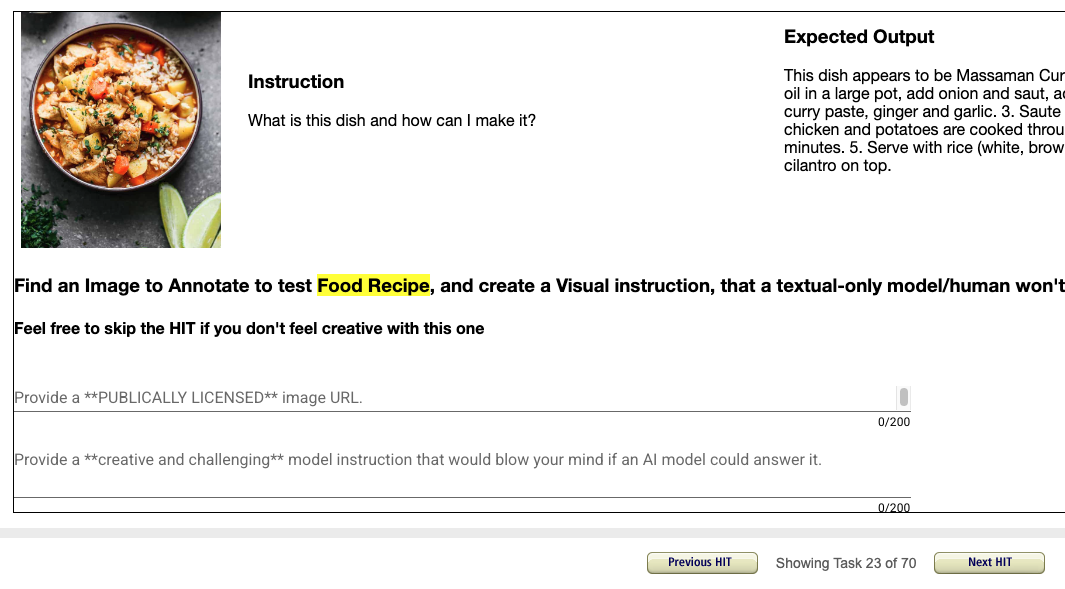}
    \caption{An interface that collects the image and corresponding instruction.}
    \label{fig:templates-instruction-generation}
\end{figure}

\begin{figure}[ht]
    \centering
    \includegraphics[width=\linewidth]{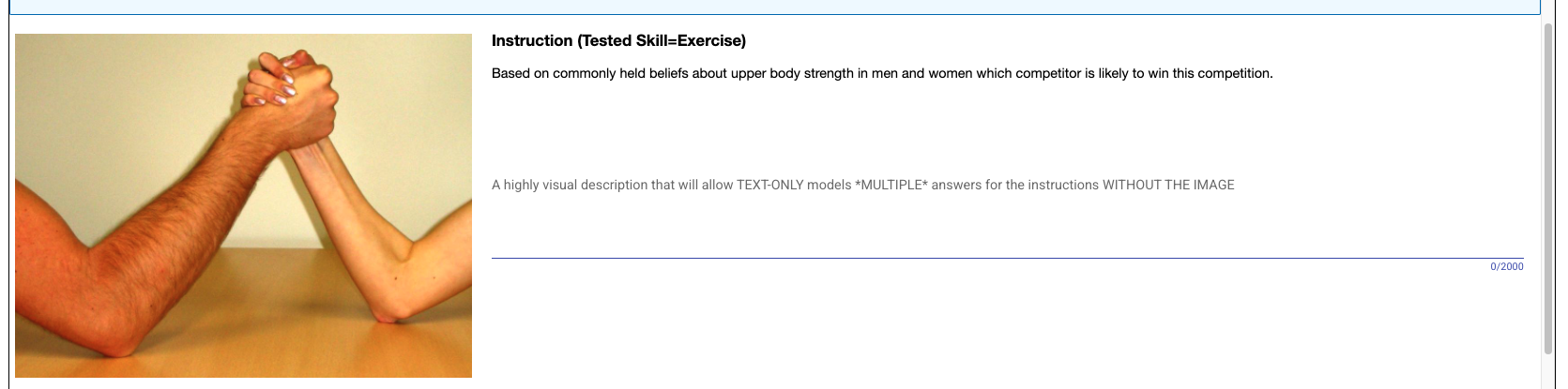}
    \caption{An interface that collects the dense caption given an image-and-instruction pair.}
    \label{fig:templates-dense-caption}
\end{figure}

\begin{figure}[ht]
    \centering
    \includegraphics[width=\linewidth]{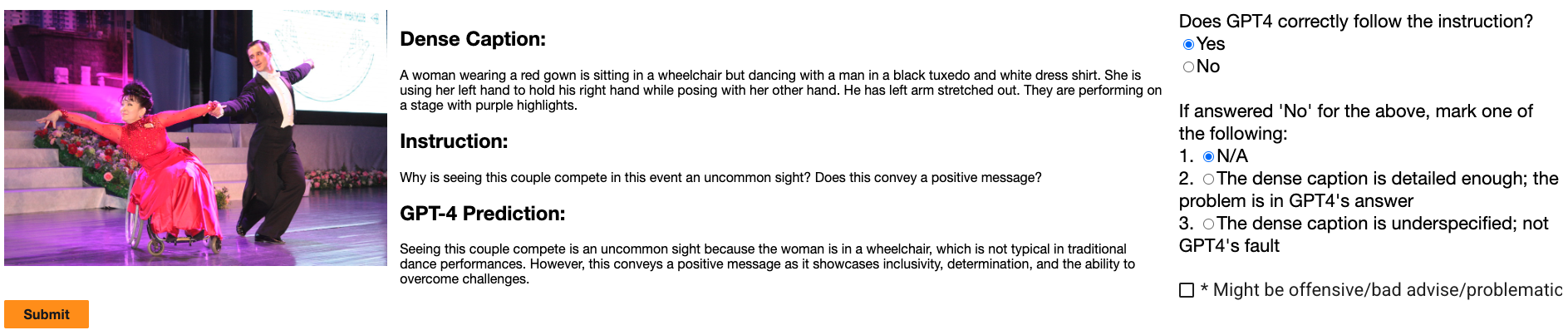}
    \caption{A sample of the model verification where the GPT4 follows the instruction correctly.}
    \label{fig:templates-human-veri-pos}
\end{figure}

\begin{figure}[ht]
    \centering
    \includegraphics[width=\linewidth]{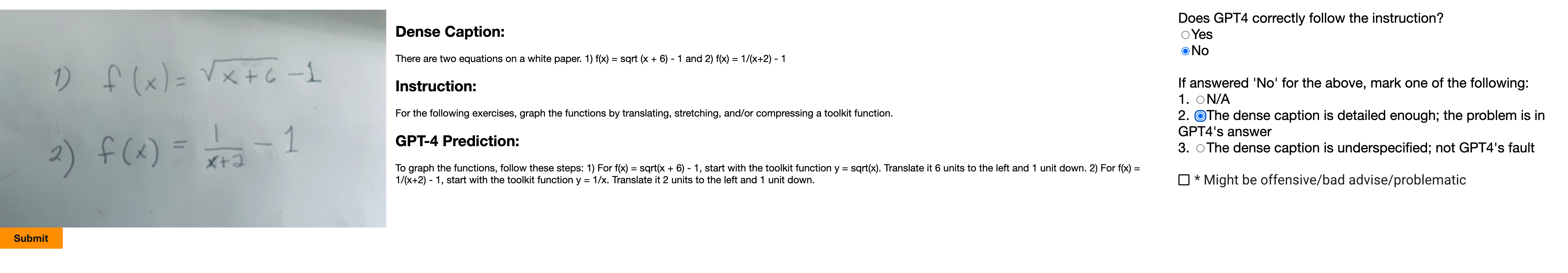}
    \caption{A sample of the model verification where the GPT4 failed to follow the instruction due to its incorrect reasoning.}
    \label{fig:templates-human-veri-neg}
\end{figure}

\begin{figure}[ht]
    \centering
    \includegraphics[width=\linewidth]{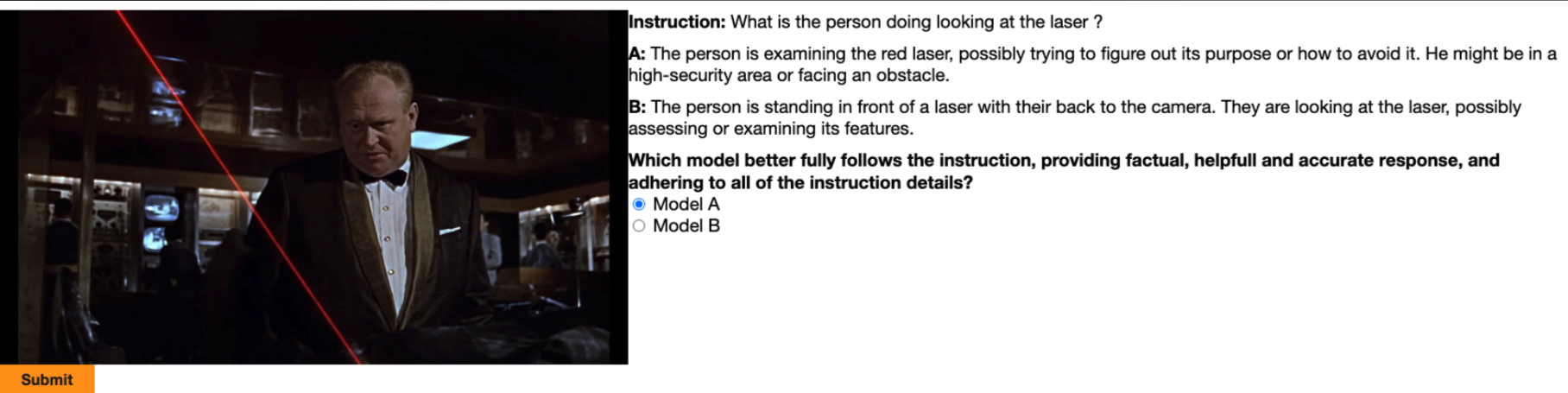}
    \caption{An interface that collects the feedback of the model rating.}
    \label{fig:templates-model-rating}
\end{figure}

In this section, we provide the templates we used to collect human annotations for the instruction generation (\Cref{fig:templates-instruction-generation}), the dense caption generation (\Cref{fig:templates-dense-caption}), the model verification (\Cref{fig:templates-human-veri-pos} and \Cref{fig:templates-human-veri-neg}), and the model rating (\Cref{fig:templates-model-rating}). 

\section{Existing Datasets incorporated in \benchmarkname}
\label{sec:existing_datasets_list}
In Table~\ref{table:existing_datasets}, we listed the existing datasets that are incoprated in our \benchmarkname. Among these datasets, 15 contain a single image in each sample pair, and 10 require reasoning based on multiple images.

\begin{table}[h]
\centering
\caption{List of existing datasets in \benchmarkname, categorized as single and multiple image datasets.}
\label{table:existing_datasets}
\resizebox{\columnwidth}{!}{
\begin{tabular}{ccc}
\hline
\multirow{15}{*}{Single} & Dataset & Topic \\\hline
& VQA \cite{antol2015vqa} & Visual Question Answering\\
& VCR \cite{zellers2019recognition} & Cognition-level Visual Understanding \\
& GD-VCR \cite{yin2021broaden} & Geo-Diverse Commonsense Reasoning \\
& WHOOPS \cite{bitton2023breaking} & What Makes this Image Strange \\   
& Newyork Caption \cite{hessel2022androids} & Humor Understanding \\
& CLEVR \cite{johnson2017clevr} & Visual Question Answering\\
& Kilogram \cite{ji2022abstract} & Tangrams Identification\\   
& Harmful Memes \cite{sharma2022you} & Memes Understanding \\
& ScienceQA \cite{lu2022learn} & Science Question Answering\\
& OK-VQA \cite{marino2019ok} & Outside Knowledge Visual Question Answering\\
& AOK-VQA \cite{AOKVQA} & Outside Knowledge Visual Question\\
& AOK-VQA \cite{AOKVQA} & Question Generation\\
& VizWiz \cite{bigham2010vizwiz} & Visual Question Answering\\
& GQA \cite{hudson2019gqa} & Visual Question Answering on Scene Graphs\\
& TextCaps \cite{sidorov2020textcaps} & Visual Question Answering on Texts\\
\hline
\multirow{10}{*}{Multiple} & Robust Change Captioning \cite{park2019robust} & Describing What has Change in a Scene\\
& NLVR2 \cite{suhr2018corpus} & Testing Visual Language Bias\\
& ImageCoDE \cite{krojer2022image} & Image Retrieval\\
& Spot-the-Diff \cite{jhamtani2018learning} & Identifying Differences\\
& VASR \cite{bitton2022vasr} & Visual Analogies\\
& WinoGavil \cite{bitton2022winogavil} & Visual Associations\\
& IRFL (Metaphor) \cite{yosef2023irfl} & Figurative Speech Understanding\\
& IRFL (Idioms) \cite{yosef2023irfl} & Figurative Speech Understanding\\
& IconQA \cite{lu2021iconqa} & Abstract Diagram Understanding\\
& Pick-a-Pic\cite{kirstain2023pick} & Text-to-Image User Preferences
\\\hline
\end{tabular}}
\end{table}

\begin{table}[h]
\centering
\caption{List of skills and existing datasets in \benchmarkname}
\label{tab:skills}
\resizebox{\textwidth}{!}{%
\begin{tabular}{|p{11cm}|}
\hline
`scienceqa', `ocr math', `recognition', `okvqa', `house plan understanding', `nlvr2', `gardening tips', `textcaps', `architectural styles', `dressing sense', `winoground', `food recipe', `paper folding', `whoops', `spot the diff', `winogavil', `imagecode', `exercise', `art knowledge', `gqa', `physical knowledge', `contextual knowledge of events', `home renovation', `aokvqa', `animals', `vasr', `counting', `board games', `solving geometry problems', `who to call?', `clevr', `building materials', `hazard identification', `pickapick', `astronomy', `figurative speech explanation', `write a story', `gestures understanding', `newyork', `cultural knowledge', `aokvqg', `traffic sign identification', `pop culture', `fashion products', `harmful memes', `write a poem', `vizwiz', `guesstimate of capacity', `location understanding', `graph reasoning', `vqa', `game playing', `differently abled', `chemical identification', `history knowledge', `climate and weather understanding', `irfl metaphor', `human emotion recognition', `medical', `gd vcr', `vcr', `technical support', `catchy titles', `kilogram', `anagrams', `color', `tour guide', `directions', `irfl idiom', `rcc'
\\\hline
\end{tabular}}
\end{table}

\section{\elo Rating}
\label{app:elo_rating}

For many years, the \elo rating has been popular in ranking players in zero-sum games such as chess \cite{elo1967proposed}. Recently, it has been adopted to rate large language models (LLMs) against each other on the user instructions. In this work, we adopt the same strategy to rank a set of instruction-following vision-language models, that can grow dynamically with further advances in the field.

Given two multimodal chatbots $\mathcal{C}_a$ and $\mathcal{C}_b$ with their absolute \elo rating $\mathcal{R}_a$ and $\mathcal{R}_b$, respectively. Simply put, the probability of $\mathcal{C}_a$ winning over $\mathcal{C}_b$ in a head-to-head battle is given by:

\begin{equation}
    P(\mathcal{C}_a \text{ wins over } \mathcal{C}_b) = \frac{1}{1 + 10 ^ {(\mathcal{R}_a - \mathcal{R}_b)/400}}
\end{equation}

In practice, calculating the \elo rating requires us to set hyperparameters to decide the weightage for each win and loss in a head-to-head battle between two models. In our work, we use the open implementation of \elo for LLMs by FastChat at \url{https://github.com/lm-sys/FastChat/blob/main/fastchat/serve/monitor/elo_analysis.py}.


\label{multi-auto-scores}

\section{GPT-4 Pairwise Evaluation Prompts}
\label{app:prompts}

The specific prompts we use to extract pairwise judgements from our language model are provided in Table~\ref{tab:full_prompt_example_gpt4_reference_free} (reference-free version) and Table~\ref{tab:full_prompt_example_gpt4_reference_backed} (reference-backed version). When applied to GPT-4 \cite{gpt4}, these prompts usually solicit a definitive pairwise response by the model. But, in some cases, the model either produces a pairwise judgement in an unexpected format, or, refuses to issue a judgement at all. For cases like these, we issue an additional query to ChatGPT to extract an answer (or decide there is no answer) using an additional prompt, given in Table~\ref{tab:full_prompt_example_chatgpt_answer_extraction}. If after this step there is still no definitive pairwise judgment, we call the result a tie.

\begin{figure*}
\centering
\resizebox{.95\linewidth}{!}{

\begin{tabular}{p{1.3\linewidth}}

\toprule
\underline{\texttt{system prompt} (human authored)}\\
\vspace{.01cm}
You are ImageTaskEvaluationGPT, an expert language model at judging whether or not a response adequately addresses an instruction in the context of an image. More specifically, you will be given the following:\\\\

1. An image context: This will describe the contents of an image with sufficient detail to address the instruction.\\
2. An instruction: This is a question, an imperative request, or something similar about the image which requires a response.\\
3. Two responses, response A and response B: These two responses attempt to address the instruction in the context of the image.\\

Your job is to judge whether response A or response B better. A and B are randomly ordered.\\\\

Some things to remember:\\\\

- Even though you are just a language model, the image description will be sufficiently detailed so that your judgements can be accurate.\\
- You are capable of judging response quality, accounting for important factors like correctness, relevance, fluency, specificity, etc.\\
- You think step-by-step, but ultimately respond with "Response A" or "Response B"\\
\midrule
\underline{\texttt{user} (human authored)}\\
\vspace{.01cm}
I will describe an image to you, and provide an instruction. Then, I will give you two candidate responses that address the instruction in the context of the image: these will be labelled "Response A" and "Response B". Your job is to first reason step-by-step about which response is best in terms of accuracy, specificity, fluency, etc. After reasoning step-by-step and comparing the pros/cons of each response, in the end, respond with "Overall, Response X is better." where X is either A or B. \\\\

\underline{\texttt{assistant} (human authored)}\\
\vspace{.01cm}
Sure, please provide the image context, the instruction, and the two candidate responses, Response A and Response B. Then, I will think step-by-step and provide my ultimate judgement as to which response is better. \\\\
\underline{\texttt{user} (from human-authored template)}\\
\vspace{.01cm}
OK. Here is the image description, the instruction, and the two response options, Response A and Response B.
Image context: A skier is wearing a green jacket, burgundy pants, white shoes, white gloves, and a gray beanie. There are people sitting on the background with their snowboards.\\\\

Instruction:
Answer the following question based on the provided image: Are the trousers the same color as the shoes?\\\\

Response A: no\\\\

Response B: No, the trousers are not the same color as the shoes. The trousers are burgundy, while the shoes are white.\\\\

Think step-by-step and finish your response with "Overall, Response X is better." where X is either A or B. \\\\
\underline{\texttt{assistant} (machine generated; used to derive the prediction)}\\
\vspace{.01cm}
Step 1: Identify the key information from the image description.\\\\

The skier is wearing:\\
- Burgundy pants\\
- White shoes\\\\

Step 2: Assess the accuracy of each response in addressing the instruction.\\\\

Both Response A and Response B convey that the trousers and shoes are not of the same color.\\\\

Step 3: Compare specificity and detail of each response.\\\\

Response A is concise and provides a straightforward answer, but lacks detail.\\
Response B provides a clear, comprehensive answer that includes the color of both the trousers and shoes.\\\\

Step 4: Consider the fluency and correctness.\\\\

Both responses are fluent and grammatically correct.\\\\

Overall, Response B is better. \\

\bottomrule

\end{tabular} }

\caption{The prompt we use for reference-free evaluation, along with a sample instance/completion from GPT-4 \cite{gpt4}.}
\label{tab:full_prompt_example_gpt4_reference_free}

\end{figure*}

\begin{figure*}
\centering
\resizebox{.95\linewidth}{!}{

\begin{tabular}{p{1.3\linewidth}}

\toprule
\underline{\texttt{system prompt} (human authored)}\\
\vspace{.01cm}
You are ImageTaskEvaluationGPT, an expert language model at judging whether or not a response adequately addresses an instruction in the context of an image. More specifically, you will be given the following:\\\\

1. An image context: This will describe the contents of an image with sufficient detail to address the instruction.\\
2. An instruction: This is a question, an imparative request, or something similar about the image which requires a response.\\
3. A reference output: This is a high-quality example output that humans have judged to be an accurate response for the input instruction.\\
4. Two responses, response A and response B: These two responses attempt to address the instruction in the context of the image.\\\\

Your job is to judge whether response A or response B better. A and B are randomly ordered. Some things to remember:\\

- Even though you are just a language model, the image description will be sufficiently detailed so that your judgements can be accurate.\\
- Take the high-quality reference into account when making your judgements, but remember: some instructions are more open-ended than others, so for those cases, a high quality response can differ from the reference.\\
- You are capable of judging response quality, accounting for important factors like correctness, relevance, fluency, specificity, etc.\\
- You think step-by-step, but ultimately respond with "Response A" or "Response B"\\
\midrule
\underline{\texttt{user} (human authored)}\\
\vspace{.01cm}
I will describe an image to you, and provide an instruction. Then, I will provide a reference output which is an example of a high quality output for that instruction in the context of the image. Then, I will give you two candidate responses that address the instruction in the context of the image: these will be labelled "Response A" and "Response B". Your job is to first reason step-by-step about which response is best in terms of accuracy, specificity, fluency, etc. After reasoning step-by-step and comparing the pros/cons of each response, in the end, respond with "Overall, Response X is better." where X is either A or B. \\\\

\underline{\texttt{assistant} (human authored)}\\
\vspace{.01cm}
Sure, please provide the image context, the instruction, the reference, and the two candidate responses, Response A and Response B. Then, I will think step-by-step and provide my ultimate judgement as to which response is better. \\\\
\underline{\texttt{user} (from human-authored template)}\\
\vspace{.01cm}
OK. Here is the image description, the instruction, the high-quality reference, and the two response options, Response A and Response B.\\
Image context: There is a white plate that has on it a small pizza with many topics, mostly unrecognizable, due to the pie being very well done, charred to the point of almost being burnt, and the crust blackened in many places.\\\\

Instruction:
Someone told the creator of this item that they are very talented. What is the implied meaning?\\
High quality reference: The implied meaning is that the person complimenting the creator is being sarcastic, as the pizza is overcooked and almost burnt, which is not a sign of talent.\\\\

Response A: When someone says that they are "very talented" in the context of the image, it implies that the person responsible for creating the pizza is skilled in the craft and able to create an appealing and tasty dish. The person praising the pizza's creator recognizes their culinary skills, which can be seen through the combination of ingredients and the overall look of the finished dish. The praise is a form of appreciation for their efforts and the enjoyable result they achieved.\\
Response B: It is difficult to determine the implied meaning without further context or information about the image. It appears to be a pizza that has been burnt or charred on the surface and has various toppings, including pepperoni, onion, and mushrooms. The image is shown on a white plate or surface. The implied meaning of this image could be that it is a dish that has been cooked badly or that it is a dish with a unique or unusual combination of ingredients.\\\\

Think step-by-step and finish your response with "Overall, Response X is better." where X is either A or B.\\\\

\underline{\texttt{assistant} (machine generated; used to derive the prediction)}\\
\vspace{.01cm}
Step 1: Comparing both responses with the high-quality reference, it's clear that Response A describes the pizza as well-made and appealing, while Response B is more focused on the fact that it is burnt or charred, somewhat aligning it with the reference.\\
Step 2: Evaluating specificity, Response A defines the creator as skilled, even though the image description contradicts this by stating that the pizza is overcooked. Response B, while not as specific, acknowledges the overcooked nature of the pizza.\\
Step 3: In terms of fluency, both responses are easily understood and written in coherent language.\\
Step 4: Considering accuracy, Response A is inaccurate in relation to the context and instruction because it implies a positive meaning when it should be sarcastic. Response B doesn't fully capture the sarcasm but is closer to the truth about the pizza being badly cooked.\\
Step 5: Understanding relevance, Response A is not entirely relevant due to the inaccuracy in the creator's talent, while Response B, although less direct, is more relevant by mentioning the burnt or charred nature of the pizza.\\
Overall, Response B is better.\\

\bottomrule

\end{tabular} }

\caption{The prompt we use for reference-backed evaluation, along with a sample instance/completion from GPT-4 \cite{gpt4}.}
\label{tab:full_prompt_example_gpt4_reference_backed}

\end{figure*}
\begin{figure*}
\centering
\resizebox{.95\linewidth}{!}{

\begin{tabular}{p{1.3\linewidth}}

\toprule
\underline{\texttt{system prompt} (human authored)}\\
\vspace{.01cm}
You are FinalAnswerExtractionGPT, an expert language model at extracting multiple choice answers from written out from longer explanations. You will be given several sentences describing a thought process which should eventually contain a final answer, either A or B. Your job is to extract the final answer that the explanation arrives at.\\\\

Some things to remember:\\\\

- Keep your answers short: only output "Final Answer: X" where X is A or B\\
- If the explanation doesn't mention a final choice explicitly, you can output "Unknown"\\\\

\ldots\\
(\texttt{+ 2 in-context examples} (human authored))\\
\ldots\\

\midrule

\textcolor{blue}{Example One: GPT-4 gives a definitive answer, but in the wrong format} \\\\
\underline{\texttt{user} (GPT-4's raw pairwise judgement)}\\
\vspace{.01cm}
Step 1: Compare the two responses to the instruction\\

- The instruction asks if people are required to stop and requests an explanation.\\\\

Step 2: Assess Response A \\
- Response A indicates that people should stop due to the presence of a stop sign. However, it doesn't consider the fact that the sign is pressed against the bus, and the LED lights are off. \\\\

Step 3: Assess Response B \\
- Response B simply says "yes," acknowledging that people should stop, but it doesn't provide the explanation required by the instruction.\\\\

Step 4: Compare specificity, fluency, and other factors\\
- Response A is more specific and fluent but misconstrues the image description.\\
- Response B is accurate in that it acknowledges the stop sign but lacks an explanation.\\\

Overall, since neither response is fully correct, it is difficult to declare a better response. However, given the fact that Response B acknowledges the prominent stop sign, I would say that Response B is slightly better, but it should also explain how the situation indicates that people should stop.\\

Please extract the final answer from the above text. \\\\

\underline{\texttt{assistant} (machine generated; used to derive the prediction)}\\
\vspace{.01cm}
Final Answer: Response B is slightly better, but it should also explain how the situation indicates that people should stop.\\
\midrule
\textcolor{blue}{Example Two: GPT-4 refuses to answer.} \\\\

\underline{\texttt{user} (GPT-4's raw pairwise judgement)}\\\\

Step 1: Consider the image context, which states that the original thirteen colonies are shown with New Hampshire specifically highlighted in a darker green.\\\\

Step 2: Assess the accuracy of Response A, which claims that the colony shown is "Virginia." This is not accurate since the image context indicates New Hampshire being highlighted.\\\\

Step 3: Assess the accuracy of Response B, which claims that the colony shown is the Massachusetts Bay Colony. This response is also not accurate, as the image context states New Hampshire is the highlighted colony.\\\\

Step 4: In terms of accuracy, both responses are incorrect. Neither response is consistent with the image context.\\\\

Step 5: Consider other factors like fluency and specificity. Both responses seem fluent and specific in nature.\\\\

After considering these factors, none of the responses are accurate given the information provided in the image context. However, I am unable to provide an "Overall, Response X is better." judgement in this case as both responses are incorrect and neither stands out as being better than the other.\\\\

Please extract the final answer from the above text. \\\\

\underline{\texttt{assistant} (machine generated; used to derive the prediction)}\\
\vspace{.01cm}
Final Answer: Unknown.\\

\bottomrule \\

\end{tabular} }

\caption{The prompt we use to extract answers from ill-formatted GPT-4 \cite{gpt4} responses, along with two example outputs. In one case, GPT-4's pairwise judgement is given, but in a format different than requested. In the other case, GPT-4 ``defies" the prompt, and refuses to issue a judgement because both options are bad.}
\label{tab:full_prompt_example_chatgpt_answer_extraction}

\end{figure*}

\end{document}